# Tender Notice Extraction from E-papers Using Neural Network

Ashmin Bhattarai[*,1], Anuj Sedhai[1], Devraj Neupane[1], Manish Khadka[1] and Rama Bastola[1]

[1]Department of Electronic and Computer Engineering, Thapathali Campus, Institute of Engineering, Tribhuvan University, Kathmandu, Nepal
[*]Corresponding author: Ashmin Bhattarai (e-mail: tha075bct012@tcioe.edu.np, koushal.ashmin@gmail.com).

**ABSTRACT**
Tender notices are usually sought by most of the companies at regular intervals as a means for obtaining the contracts of various projects. These notices consist of all the required information like description of the work, period of construction, estimated amount of project, etc. In the context of Nepal, tender notices are usually published in national as well as local newspapers. The interested bidders should search all the related tender notices in newspapers. However, it is very tedious for these companies to manually search tender notices in every newspaper and figure out which bid is best suited for them. This project is built with the purpose of solving this tedious task of manually searching the tender notices. Initially, the newspapers are downloaded in PDF format using the selenium library of python. After downloading the newspapers, the e-papers are scanned and tender notices are automatically extracted using a neural network. For extraction purposes, different architectures of CNN namely ResNet, GoogleNet and Xception are used and a model with highest performance has been implemented. Finally, these extracted notices are then published on the website and are accessible to the users. This project is helpful for construction companies as well as contractors assuring quality and efficiency. This project has great application in the field of competitive bidding as well as managing them in a systematic manner.

## INTRODUCTION

An official invitation or offer to supply goods or execute a particular task, projects is known as tender. Tendering is the process of invitations for various projects by the government or other private institutions mentioning different criteria. In the context of Nepal, tender is generally published in the newspaper. But as the new generation is becoming more connected to the digital world, the tender notice published in the Newspaper becomes less effective.

Artificial Intelligence (AI) is the branch of computer science which deals with helping machines to find solutions to complex problems in a more human-like fashion. AI can be found being widely used in the field of medicine, computer science, transportation and so on. A convolutional neural network (CNN) is a type of artificial neural network used in image processing and recognition that is specifically designed to process pixel data. CNNs are effective image processing, artificial intelligence that use deep learning to perform both generative and descriptive tasks, often using machine vision that includes image and video recognition. It consists of Convolutional Layer, Activation Layer, Pooling Layer, and Fully Connected Layer, and these all layers are interconnected so that CNNs can effectively process and perceive data in order to classify images.

A CNN architecture is formed by stacking different layers that can transform the input volume into an output volume (e.g., holding the class scores) through a differentiable function. Three different CNN architectures ResNet [1], GoogleNet [2] and Xception [3] are formed with different combinations of layers.

Since the tender notices may be in textual or image format and in order to maintain uniformity, a common approach should be adopted while dealing with this problem. This is the reason why we opted to convert all the pages of newspapers into individual images so that notice in both formats can be extracted. It can be applicable for contractors, construction companies, real estate companies, etc. Government institutions and working bodies can directly publish their tender notices in the website rather than publishing in newspapers and reduce their cost.

The first section of this paper provides general information about the project including objectives, problem definition,

scope and application of the project. Likewise, the second section provides the review of the related works and the third section is about methodology. Furthermore, the fourth and fifth sections are about result and conclusion respectively.

## RELATED WORKS

Manuel J. García Rodríguez, et. al. [4] developed a bidders recommender system for public procurement auctions which could recommend potential bidders using machine learning, particularly random forest classifiers. In this project, the winning bidding company was within the recommended companies' group, from 24% to 38% of the tenders, according to different test conditions and scenarios.

Wei-Ta Chu, et. al. [5] proposed advertisement detection and segmentation where the system detected advertisement candidates based on a connected components method and used CNN with SVM to classify advertisement from the rest of the contents. However, due to insufficient datasets, they achieved average 85.54% accuracy using only CNN, 84.65% using only semantics and 94.04% using both.

Pooja Jain, et. al. [6] from Punjab university, Department of Science and application developed a convolutional neural network-based advertisement classification models for online newspapers that can classify advertisement images from English newspapers into four predefined categories. Initially, they used a simple CNN-model for advertisement classification purposes which in turn gave an accuracy of 65%. Similarly, they used ResNet50 architecture along with default hyperparameters and achieved accuracy of 68%. Finally, implementation of ResNet50 architecture with Fine Tuned hyperparameters yielded an accuracy of 74%.

Kim Sungyoung, et. al. [7] proposed a method for extracting interesting objects from complex backgrounds. In this method, a core object region was selected as a region a lot of pixels of which had the significant color, and then it was grown by iteratively merging its neighbor regions and ignoring background regions. But in this project some inaccuracies occurred because of the under-extraction or over-extension of wrong significant regions.

An Tien Vo, et. al. [8] proposed image classification model that can be applied for identifying the display of online advertisement. They used a convolutional neural network with two parameters (n, m) where n is a number of layers and m is the number of filters in the Conv layer. However, using Convolutional Neural Network only extracts image advertisements only but advertisements might be published in textual form as well. In such scenarios, this model might be inefficient.

## METHODOLOGY

There are no extensive datasets to train the model therefore we manually created the dataset from newspaper notices, Bolpatra Nepal and google photos as well.

### Models Building

#### ResNet Model

First of all, a class called 'Residual Unit' was defined. The constructor of this class will take list containing number of filters, strides and activation function as parameters. Inside constructor, Residual Unit was designed. Similarly, another method call() was also defined inside this class which will implements Residual Unit. This method will take input block as parameter as return the result after convolution and concatenation of input block. There are also two others methods: - build_graph() and get_config() which were used to generate pictorial form model and configure all parameters used. Once Residual Unit class was made, it was used to make whole ResNet model. This model will take input images of size 224 x 224 in grayscale. Then a convolution layer followed by BN layer and ReLU activation function was used. For convolution layer, convo2D() method inside keras Layer class was used. Similarly, BatchNormalization() and Activation() method were used for BN and activation layers. 'ReLU' was passed as argument for Activation() method. Now, max pooling was done using MaxPool2D() method. 41 Now four times Residual Unit was stacked. And after last Residual Unit averagepooling was done using GlobalAveragePooling2D() method. At last, there were three dense layers with 32, 2 and 1 neurons

#### GoogLeNet Models

Like the Residual Unit, Inception class was created for this model and this class has all the methods in the Residual Unit. The only difference in this class is that, in its constructor, the Inception module was designed. To create the Inception module Convo2D(), BatchNormalization(), Activation(), MaxPool2D() and Concatenate() methods were used.
After building Inception class, it was ready to build GoogLeNet model. Similar to ResNet model, this model will also take input images, applies convolution, batch normalization, activation and max pooling. Now one inception module was used followed by a max pool layer. Then, there are 3 more Inception modules and after 3 inception modules, there was one max pool layer. Furthermore, one more inception module was kept and followed by global average pooling. At last, there are 3 dense layers with 32, 2 and 1 neurons. Xception Model As discussed earlier this model combines idea of ResNet and GoogLeNet model, here also a class called 'Xception Unit' was created. Its constructor looks very similar to constructor of Residual Unit class where Convo2D() method was replaced by SeperableConvo2D(). But two additional parameters 'isEntryExit' and 'isFirst' can be seen which are False by default. If both 'isEntryExit' and 'isFirst' was True then first separable convolution layer will be remover form the module. If only 'isEntryExit' is True then a max pool layer is added to very last to this module also a skip connection layer will be created. Now like other models Xception model will take input, apply separable convolution followed by batch normalization and activation layer. Unlike other two models max pool layer was not used here because it has been used inside Xception Unit class. Hence, a Xception Unit was added with both 'isEntryExit' and 'isFirst' equal to True. This layer also called Entry flow unit. Similarly, three middle flow units were placed (i.e. Xception Unit with default arguments). Now an exit flow layer was places followed by a separable convolution, batch normalization, activation and flatten layers. At last there was one dense layer with 2 neurons. Models Training Now models were first trained up to 50 epochs using fit() method. Then all models were evaluated on test dataset. Furthermore, all three models were trained up to 100 epochs. The results were compared and one model was chosen as the final model.

**Downloading E-papers**
First a request is made to the gorkhapatra website with the help of a web driver instantiated earlier. It navigates to the homepage of that website. Next a search to an element is made by the giving xpath of that element to the find element method of the web driver. After finding the element a click operation to that element is made to navigate to the location point by that link. Next the page source of this new page is fetched with the help of the page source method of the web driver on which an element with class attribute set to "pdf" is identified. On this element a link to the actual pdf file is present. By making a download request to this link the web driver starts to download that pdf. Until the pdf is downloaded a repeated check to the download location is made to identify whether the file is downloaded completely or not.

**PDF to Image Conversion**
Files with ".PDF" extension are opened using fitz.open() by providing the path of the file which will create an object of a class Document. The provided parameters are the constructor parameter of the class Document. After loading the PDF file, iterate through 0 to the number of pages of that file. During each iteration, a Page object is created using fitz.doc.load_page(page_number). Page is a class representing a document page. Now to convert that page into image format, pixmap is obtained by using fitz.page.get_pixmap(). Pixmap ("Pixel Map") represents plane rectangular sets of pixels. Each pixel is described by a number of bytes ("components") defining its color, plus an optional alpha byte defining its transparency. To control the quality of the final image, DPI parameters can be passed to fitz.page.get_pixmap() which determines the number of pixels present in one inch in both x and y direction. Then the pixmap is saved as an image file using pixmap.save(). This is repeated for every page of the PDF.

**Extraction of sub images**
First the image is converted to grayscale and thresholding is applied. The rectangular contours are determined using OpenCV built-in function. The contour is checked for the dimension threshold and checked if it has any child contour inside of it. If it does, then it is also further checked for dimension. Bounding coordinate of contour meeting the determined threshold is calculated and image is cropped using those coordinates from original image.

**Image Padding and Resizing**
To convert images into required format several library functions of OpenCV were used. First image was read using cv2.imread() function which takes image name as argument and returns image in NumPy array [9]. Also, another argument can be passed to specify color mode in which the image will be read. This parameter was set to default hence, the image was read in RGB format. Now a null matrix was created using the np.full() function of the NumPy array, where image width and height are passed as arguments to specify the shape of the matrix. Furthermore, few other parameters like integer type were also passed. Now the original image was copied inside a new null matrix. After that, new image was converted to 224x224 size using cv2.resize()

**Implementation of CNN Model**
First, all the images from a folder are read one by one with the help of cv2.imread function. Then, images are resized to 224x224 dimensions. After that, all the images are converted into grayscale format with the help of cv2.cvtColor. Now the grayscale image is converted into NumPy Array. Finally, all the images are appended to a single NumPy
array. Now with the help of keras API [10] and object of Xception model is instantiated and the NumPy array of images prepared earlier is fed to this model to predict whether the image is notices or not. After predicting all the images, model sends back a two-dimensional array consisting of image and its label. Now, label for individual image is compared. If the label of an image is positive, then it is stored in a separate folder and those with negative labels are discarded.

**Implementation of OCR**
First image is fed into the Tesseract [11] OCR engine. It will return a pandas data frame with different columns like level, page_num, block_num, par_num, line_num, word_num,
left, top, width, height, conf, text. Then the file which consists of all the keywords related to tender notice are loaded. Then the list of text is detected using OCR and the list of words are read from the dictionary file to "set" data type. Then these two sets are intersected and the number of intersections is counted. If the number of common words is greater than the predefined number, it will be declared as tender.

**RESULT ANALYSIS**
After training all three models up to 50 epochs following results were obtained.

**Training Up to 50 Epochs**

**ResNet Model**

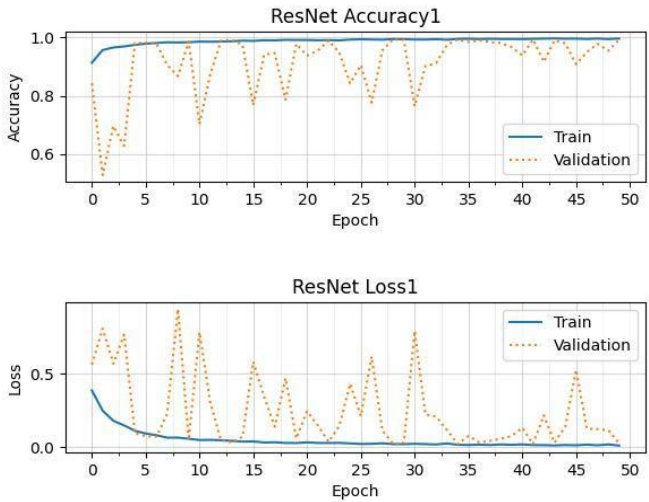

*Figure 1: ResNet Accuracy and Loss vs Epochs (50)*

Firstly, ResNet Model was trained and an accuracy of 0.9949 was achieved. However, a loss of 1.96 was also determined during the process. Likewise, precision of 0.9945, recall of 0.9955 and F1 score of 0.9950 were obtained after the completion of training.

**GoogleNet Model**

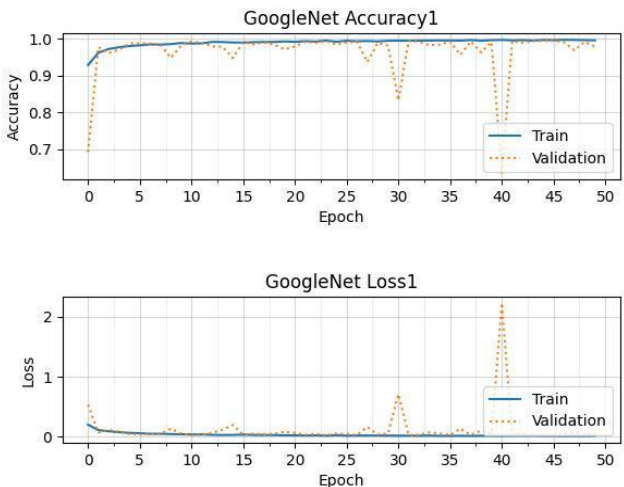

*Figure 2: GoogleNet Accuracy and Loss vs Epochs (50)*

Similarly, GoogLeNet Model was trained and an accuracy of 0.9959 was achieved. However, a loss of 1.43 was also determined during the process. On the other hand, precision of 0.9955, recall of 0.9965 and F1 score of 0.9960 were obtained after the completion of training.

**Xception Model**

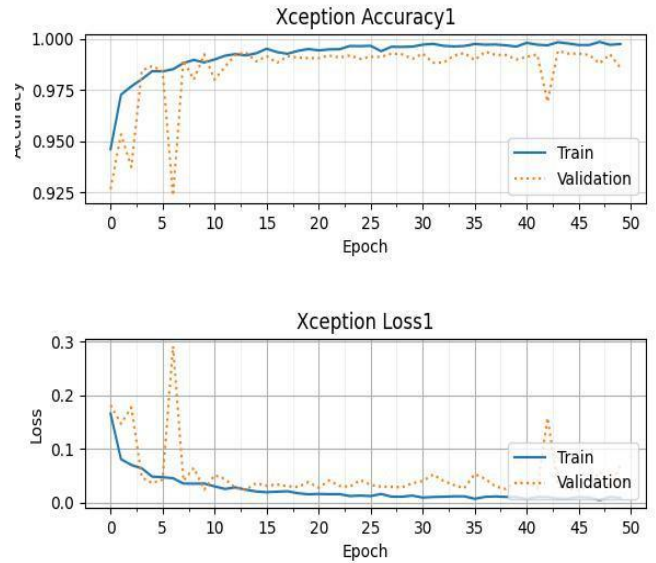

*Figure 3: Xception Accuracy and Loss vs Epochs (50)*

Finally, the Xception Model was trained and an accuracy of 0.9926 was achieved. However, a loss of 2.5% was also determined during the process. Similarly, precision of 0.9964, recall of 0.9984 and F1 score of 0.9974 were obtained after training of the Xception model.

Training all three models up to 50 epochs gives the following results.

*Table 1: Results obtained while training models up to 50 epochs*

| Models | Trained up to 50 epochs | | | |
|---|---|---|---|---|
| | Accuracy | Precision | Recall | F1 score |
| ResNet | 0.9949 | 0.9945 | 0.9955 | 0.9950 |
| GoogLeNet | 0.9959 | 0.9955 | 0.9965 | 0.9960 |
| Xception | 0.9926 | 0.9964 | 0.9984 | 0.9974 |

**Training Up to 100 Epochs**
Since ResNet and GoogLeNet models were not performing well, this might be due to training up to 50 epochs only. So, all three models were again trained up to 100 epochs and following results were obtained.

## ResNet Model

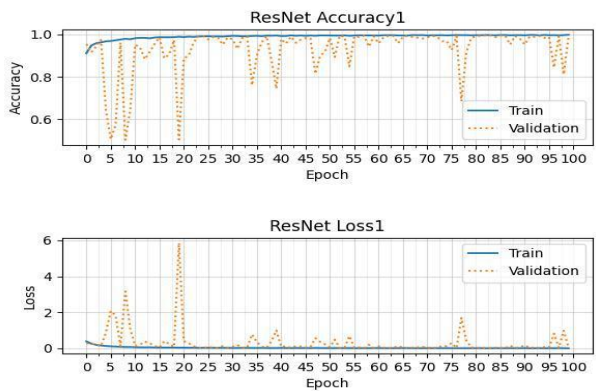

*Figure 4: ResNet Accuracy and Loss vs Epochs (100)*

ResNet gave accuracy of 0.9926 with loss of 0.0252. Similarly, precision, recall and F1 score were 0.9949, 0.9904 and 0.9927 respectively.

## GoogLeNet Model

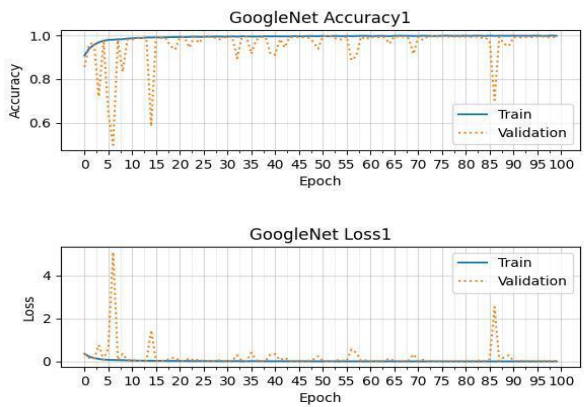

*Figure 5: GoogleNet Accuracy and Loss vs Epochs (100)*

GoogLeNet gave accuracy of 0.9947 with loss of 0.0171. Similarly, precision, recall and F1 score were 0.9945, 0.9950 and 0.9947 respectively.

## Xception Model

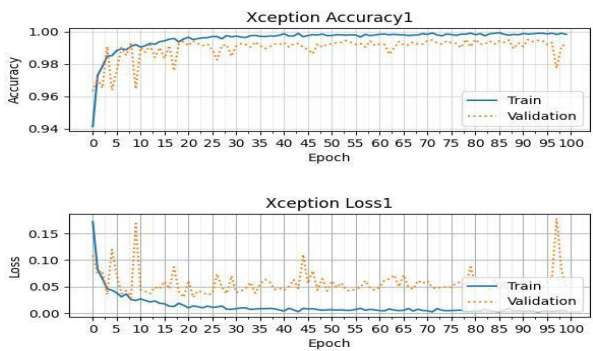

*Figure 6: Xception Accuracy and Loss vs Epochs (100)*

Xception gave accuracy of 0.9919 with loss of 0.0318. Similarly, precision, recall and F1 score were 0.99946, 0.9963 and 0.9954 respectively.

Training all three models up to 50 epochs gives the following results.

*Table 2: Results obtained while training models up to 100 epochs*

| Models | Trained up to 100 epochs | | | |
|---|---|---|---|---|
| | Accuracy | Precision | Recall | F1 score |
| ResNet | 0.9926 | 0.9949 | 0.9904 | 0.9927 |
| GoogLeNet | 0.9947 | 0.9945 | 0.9950 | 0.9947 |
| Xception | 0.9919 | 0.9946 | 0.9963 | 0.9954 |

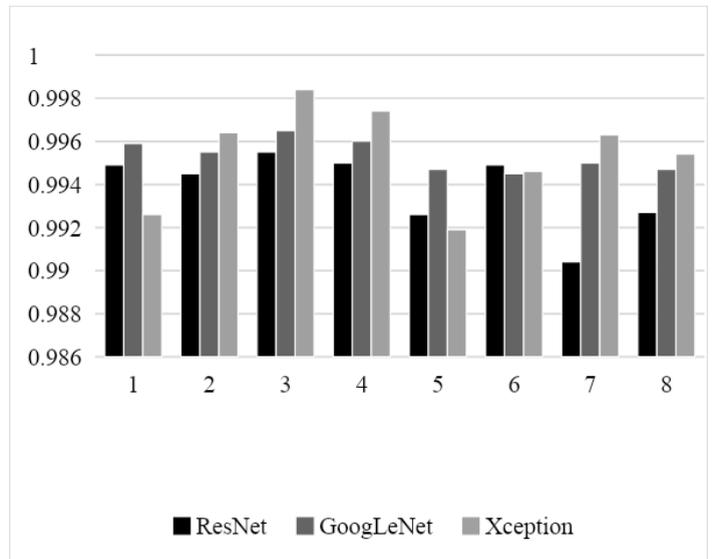

*Figure 7: Comparison between models*

From the above results, it can be seen that performance of the ResNet model in both training instances were more fluctuating with respect to epoch. Heavy decrease in accuracy with a big spike in loss was obtained. Similarly, precision and recall were also very less, which resulted in lower F1 score.

On the other hand, accuracy and loss in the GoogLeNet model is more promising than the ResNet model. But precision, recall and F1 score was not so satisfactory.

However, in the case of the Xception model it can be seen that accuracy is gradually increasing with decrease in loss. Here, larger spikes were not seen as compared to previous models. Also Recall is higher which leads to better classification. Similarly, other performance parameters like

Precision, Recall and F1 score were also seen to be better than previous models.

But while training models up to 100 epochs F1 scores had decreased for all three models. This might indicate models were overfitting the training data.
Analyzing different performance parameters, it can be seen that the Xception model trained up to 50 epochs has performed better with given datasets. Hence the Xception model will be used to classify images.

**CONCLUSION**
An image processing-based approach to segment and extract tender notices from the given source of newspapers was presented. For segmentation, rectangular edge detection technique was implemented. Similarly, in order to extract notices, different architectures of CNN (ResNet, GoogleNet and Xception) were tested and architecture with the highest performance was implemented. Among the three tested architectures, performance parameters of Xception was consistent and hence was selected as the final architecture for the classification of notices. Furthermore, the classified notices were filtered out by implementing OCR. Hence, the objective to extract tender notices was completed with high performance having recall and F1 score of 0.9984 and 0.9974 respectively.

Currently, mostly tender notices based on rectangular edges were detected. In future work, natural language processing can be used to extend this procedure. Likewise, segregation of tender notices into different categories can be considered as future enhancement. Tender notices from local newspapers can also be published.